\documentclass[letterpaper]{article} 
\usepackage{aaai2026}  
\usepackage{times}  
\usepackage{helvet}  
\usepackage{courier}  
\usepackage[hyphens]{url}  
\usepackage{graphicx} 
\usepackage{natbib}  
\usepackage{caption} 
\usepackage{algorithm}
\usepackage{algorithmic}
\usepackage{amssymb,xcolor, amsmath}
\usepackage{soul,color}
\usepackage{multicol}
\usepackage{booktabs, array, makecell}
 
\DeclareMathOperator*{\argmax}{arg\,max}
\DeclareMathOperator*{\argmin}{\arg\!\min}

\newcommand{\prior}{\text{\texttt{EVaLP}}}
\newcommand{\snis}{\text{\texttt{EVaLP+SIR}}}
\newcommand{\E}{\mathbb{E}}
\newcommand{\R}{\mathcal{R}}
\newcommand{\N}{\mathcal{N}}
\newcommand{\EL}{\boldsymbol{\mathcal{L}}}

\newcommand{\KL}{\text{KL}}

\newcommand{\mathfnt}{8} 
\newcommand{\bse}{9.6}  
\usepackage{amsthm}
\newtheorem{proposition}{Proposition}

\usepackage{algorithm}
\usepackage{algorithmic}

\usepackage{newfloat}
\usepackage{listings}
\DeclareCaptionStyle{ruled}{labelfont=normalfont,labelsep=colon,strut=off} 
\floatstyle{ruled}
\newfloat{listing}{tb}{lst}{}
\floatname{listing}{Listing}
\title{Learning Energy-based Variational Latent Prior for VAEs}
\author {
    Debottam Dutta,
    Chaitanya Amballa,
    Zhongweiyang Xu,
    Yu-Lin Wei,
    Romit Roy Choudhury
}
\affiliations {
    University of Illinois Urbana-Champaign\\

}

\usepackage{bibentry}

\begin{document}
\maketitle
\begin{abstract}
Variational Auto-Encoders (VAEs) are known to generate blurry and inconsistent samples. 
One reason for this is the ``prior hole'' problem. 
A \textit{prior hole} refers to regions that have high probability under the VAE's prior but low probability under the VAE's posterior. 
This means that during data generation, high probability samples from the prior could have low probability under the posterior, resulting in poor quality data.  
Ideally, a prior needs to be flexible enough to match the posterior while retaining the ability to generate samples fast.
Generative models continue to address this tradeoff.

This paper proposes to model the prior as an energy-based model (EBM).
While EBMs are known to offer the flexibility to match posteriors (and also improving the ELBO), they are traditionally slow in sample generation due to their dependency on MCMC methods.
Our key idea is to bring a variational approach to tackle the normalization constant in EBMs, thus bypassing the expensive MCMC approaches. 
The variational form can be approximated with a sampler network, and we show that such an approach to training priors can be formulated as an alternating optimization problem. 
Moreover, the same sampler reduces to an implicit variational prior during generation, providing efficient and fast sampling. 
We compare our {\em Energy-based Variational Latent Prior} ({\prior}) method to multiple SOTA baselines and show improvements in image generation quality, reduced prior holes, and better sampling efficiency.
\end{abstract}
\vspace{-10pt}
\section{Introduction}
Variational Autoencoders (VAE) are decoder-based latent variable models \cite{DBLP:journals/corr/KingmaW13} that have gained significant popularity in diverse applications, including image/speech generation \cite{brock2019largescalegantraining,DBLP:journals/corr/abs-2005-00341}, image captioning \cite{aneja_caption, aneja_caption_seq, caption_cvae}, and representation learning \cite{som_vae, vqvae}.  
A VAE is trained by maximizing an Evidence Lower Bound (ELBO); 
this objective balances between the VAE's reconstruction quality (via a reconstruction loss) and sample generation quality (by maximally aligning the aggregate posterior distribution to the fixed prior distribution). 
During generation, the fixed prior is used as a proxy for the aggregate posterior to generate samples from the latent space. 
These samples are passed through the decoder to generate new data.

\begin{figure}[htb] 
\centering
\includegraphics[width=0.6\columnwidth]{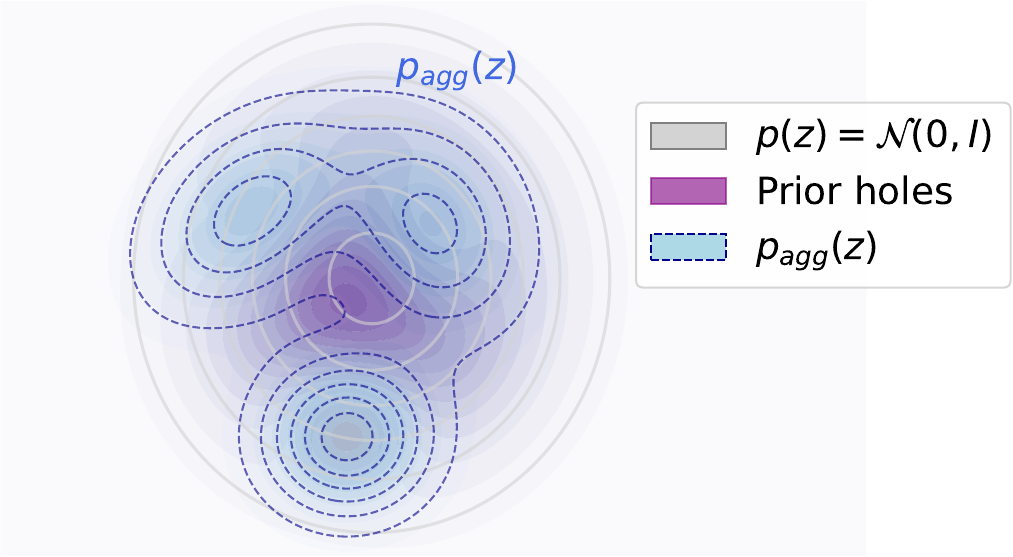}
\vspace{-5pt}
\caption{Prior hole in a 2-dimesional latent $z$ space of VAE.}
\label{fig:hole} 
\end{figure}

In vanilla VAEs, the posterior $p(z|x)$ is designed to be a Gaussian; the prior is also designed to be a fixed Gaussian $\mathcal{N}(0,I)$.
Thus, even after training has converged, there is misalignment between the aggregate posterior and the prior, causing ``prior holes''. 
Figure \ref{fig:hole} illustrates a prior hole (in 2-dimension) where regions of the latent $z$ space have high probability under the $\mathcal{N}(0,I)$ prior but low probability under the posterior. 
During generation, samples from these holes are likely but when passed through the decoder, they produce inconsistent samples.

\vspace{-3pt}

Much effort has been invested towards learning complex posterior distributions \cite{vae-posterior,nvae}. However recent research has shown that such methods alleviate the problem to some extent by improving the ELBO, but fail to fully remove the misalignment between distributions. 
Another line of attack tries to learn a better prior; this includes autoregressive priors\cite{pixel_vae}, resampled priors \cite{lars} and hierarchical prior \cite{hier_vae}.
In recent years, Energy-based models (EBM) \cite{du_ebm, latent-ebm} are becoming a popular method to learn flexible priors. 
Unfortunately, these models need complicated sampling procedures during training \cite{latent-ebm} or during generation \cite{ncp-vae,emlg, ehmlg}, which makes them computationally very expensive. 

This paper aims to address the prior hole problem by also learning a flexible EBM prior, but avoids the sampling complexity by observing that the log-normalization constant of the EBM has a variational form. 
Specifically, we formulate the VAE prior as an energy-based distribution -- an exponentially tilted Gaussian -- and use the variational form of the log-normalization constant to introduce a sampler network.  
This allows us to convert the complicated sampling process to an optimization objective, enabling the sampler to amortize the sampling process and become a variational prior after the training completes.

\begin{figure*}[htb] 
    \centering
    \vspace{-20pt}
        \includegraphics[width=0.7\textwidth]{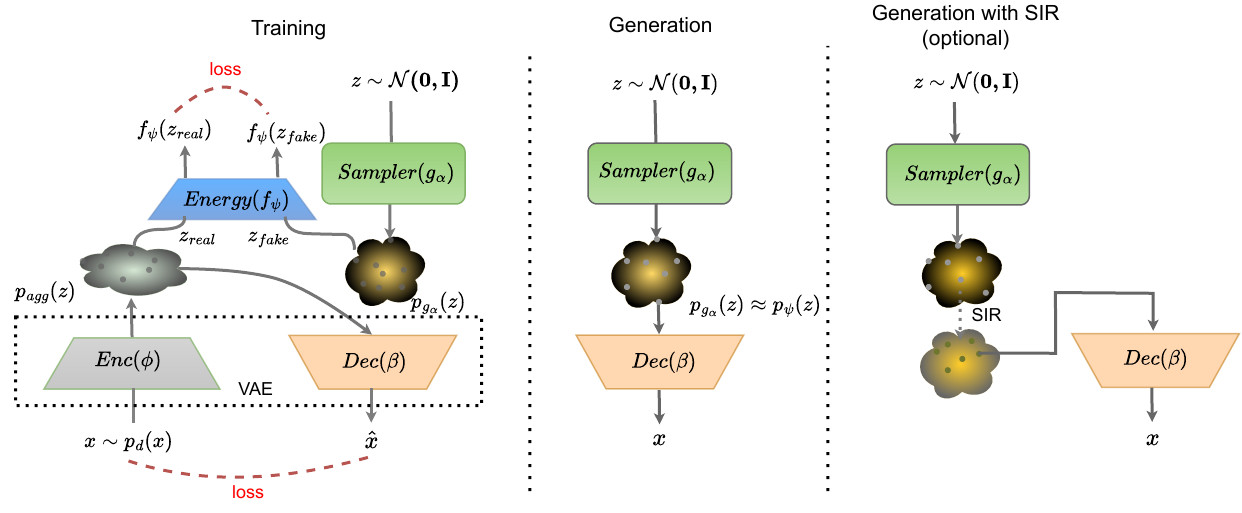}
        \vspace{-7pt}
    \caption{Proposed method: (a) Training phase that first trains a VAE, followed by a joint training of the EBM $f_\psi$ and the sampler network $g_\alpha$. (b) MCMC-free sample generation. (c) Sample generation using SIR (optional) to improve sample quality.}
    \label{fig:flowchart} 
    \vspace{-15pt}
  \end{figure*}

The variational form of the log-normalization constant leads to a max-min objective over a value function, which is an upper bound to the ELBO. Although the connection of EBM training with max-min objective is not unknown, we demonstrate that the straightforward extension of the formulation in VAE's latent space leads to some stability issues.
However, with a mild assumption on the EBM energy function, we convert the max-min problem to a more stable alternating optimization problem.   
Since we replace the log-normalization constant  with a variational form to introduce a sampler, the proposed method no longer requires MCMC sampling during either training or generation. 
Moreover, since our EBM function also learns an explicit density over the latent space, we can opt to use energy-driven sampling such as Sampling-importance resampling (SIR) \cite{eim} to further improve the generation quality. 
Figure \ref{fig:flowchart} visually summarizes our ideas. 

Our contributions are: \\ 
$\bullet$ Observing that a sampler network can realize the variational form of the log-normalization constant. \\ 
$\bullet$ Formulating an upper bound on the VAE's ELBO using the sampler network and formulating a max-min objective function; solving this through an alternating optimization approach.\\
$\bullet$ Demonstrating stable convergence, resulting in a flexible EBM prior that matches the aggregate posterior, while also permitting fast sample generation. \\
$\bullet$ Extending the approach to hierarchical VAE models and showing competitive performance with other baselines.

\section{Related Work}
\textbf{Energy Based Models (EBMs)} 
 are among the earliest deep generative models, rooted in Boltzmann machines \cite{DBLP:journals/neco/Hinton02}. 
 Despite many initial attempts to utilize EBMs for modeling complex data distributions \cite{rus_hinton, poe_hinton}, only the recent works \cite{du_ebm, ddaebm, coopnet,latent-ebm,  vaeebm, coop_desc_gen} have shown promising results. 
 Specifically, the efforts in \cite{coopnet, dual_mcmc} successfully learn an EBM in the pixel space on top of a latent variable model. On the other hand, 
 \cite{latent-ebm, adaptive_ce} have used an EBM to model a data-dependent prior for decoder-based generative models. \cite{twoflows} uses a similar method to learn a prior which a normalizing flow model in the prior and Langevin dynamics to sample from the posterior. 
Other efforts have combined EBMs with VAEs -- in pixel space \cite{vaeebm, div_triangle, xie_vae_sampler} or in latent space \cite{vae-ebm, emlg, ehmlg} -- to learn an energy model that can increase the capacity of VAEs; even they require MCMC.
  
 Toward avoiding MCMC for EBM training, 
 \cite{vera} have also shown that an online sampler network can be used to approximate the log-normalizing constant of the EBM at the expense of introducing an adversarial max-min game. 
 Along similar lines, \cite{fce} has used a normalizing flow to train an EBM using NCE \cite{nce} in the pixel space; the NCE learns a self-normalized EBM.
 This self-normalized design leads to a min-max approach, similar to vanilla-GAN~\cite{gan}. 
 
 In contrast, although we use a normalizing Flow model as a sampler for the EBM, our method is situated in the latent space of VAE. 
 We design our EBM prior with an explicit normalizing constant, leading to a max-min game like a Wasserstein-GAN (WGAN) \cite{pmlr-v70-arjovsky17a}. 
The WGAN convergence is known to be more stable and robust towards mode collapse. 
Closest to our work is NCP \cite{ncp-vae} where authors use NCE to learn an energy-based prior in the latent space of VAE. We compare our approach with NCP, in addition to other baselines. 

\textbf{Prior hole problem.}
Most approaches have tackled the mismatch between prior and aggregate posterior by learning a more complex prior. 
While \cite{lars} train a reweighting factor for the prior using truncated rejection sampling, \cite{snis} uses Sampling Importance Resampling (SIR) and \cite{hlebm_diffusion} uses Langevin Dynamics -- these kinds of methods usually incur higher computational overhead due to the use of various sampling techniques during training. While our approach also reweights the prior, it completely avoids expensive Monte Carlo sampling during training.  In a different line of approach,
Adversarial Autoencoders (AAE)\cite{aae} can generate sharp samples by training a discriminator that brings the aggregate posterior closer to the prior.
Our approach is complementary to AAE in that we train the prior using a critic/discriminator to bring it closer to the pre-trained aggregate posterior.  \\ 
Despite these similarities and existence of application of GANs in VAE's latent space \cite{ShapeGF,denovo-latent-gan}, these approaches have not been studied much from an energy-model's point of view. We establish these connections in this work by deriving the latent adversarial game from first principles.

\textbf{Two stage VAEs:}
One of the most popular two-stage VAE models is VQ-VAE. It has an auto-regressive generator that estimates the latent distribution in the 2nd stage. 
However, VQ-VAE is a deterministic model and is very slow to sample from due to this auto-regressive nature. 
Other models include 2stage-VAE~\cite{2svae}, which learns another VAE in latent space, and RAEs~\cite{rae}, which regularize autoencoders before density estimation with a Gaussian mixture model.
More recent methods such as NCP~\cite{ncp-vae} and \cite{emlg, ehmlg} are also 2-stage approaches on VAE where they learn the latent density of a hierarchical VAE by estimating an energy-based density model in the 2nd stage. While NCP needs sampling importance resampling (SIR) during generation, \cite{emlg, ehmlg} both need Langevin Dynamics to sample from the EBM prior. Our {\prior} approach learns a variational prior to completely avoid MCMC during training and generation.

\section{Background}
\subsection{Variational Auto-Encoders (VAEs)}
VAEs are latent variable models that define the data generation process as $p_\theta(x,z) = p_\beta(x|z) p_\psi(z)$ where $p_\beta(x |z )$ is the conditional distribution of the observed variable given the latent variable, and $p_\psi(z)$ is the prior distribution of the latent variable. In vanilla VAEs, the prior is a fixed Gaussian, $p_\psi(z) = \N (0, I):=p_0(z)$. The Evidence LOwer BOund (ELBO), is used as a maximization objective to train the VAE: 
\vspace{-8pt}
{\fontsize{8}{12}\selectfont
\begin{align}
    \EL(x) = \E_{q_\phi(z|x)} \big[\log p_\beta(x|z)\big] - 
    \KL\big(q_\phi(z|x) \| p_0(z)\big) \label{eq:vae_elbo}
\end{align}
} where $q_\phi(z|x)$ is a variational distribution that approximates the true posterior distribution $p_\theta(z|x)$. The variational posterior $q_\phi(z|x)$ is chosen to be another Gaussian $\N (\mu(x), \sigma(x))$, where the mean and covariance are learned from data. 
\begin{table}[h]
\vspace{-8pt}
\centering
\scriptsize
\begin{tabular}{@{}l|
    >{\centering\arraybackslash}p{1.8cm}|
    >{\centering\arraybackslash}p{1.8cm}|
    >{\centering\arraybackslash}p{1.8cm}@{}}
    \toprule
     & \makecell{Explicit density} 
     & \makecell{Faster sampling} 
     & \makecell{Efficient training} \\
    \midrule 
        VAE & $\boldsymbol{\times}$ & \textbf{\checkmark} & \textbf{\checkmark} \\
        EBM & \textbf{\checkmark} & $\boldsymbol{\times}$ & $\boldsymbol{\times}$ \\
    \bottomrule
\end{tabular}
\vspace{-4pt}
\caption{VAE versus EBM as generative models.}
\vspace{-10pt}
\label{tab:vae_vs_ebm}
\end{table}
\vspace{-4pt}
\subsection{Formulation of the Prior Hole Problem}
The VAE encoder implicitly defines a latent space distribution $\E_{p_d(x)} [q_\phi (z|x)]:= q_{agg}(z)$, called the aggregate posterior, where $p_d(x)$ is the data distribution. 
As shown in Appendix \ref{app:derivations}, 
a prior distribution maximizes the ELBO when it matches the aggregate posterior.
The distance between these two distributions controls the generation quality of the VAE. The VAE suffers from the \emph{Prior Hole Problem} when the prior fails to match the aggregate posterior. 
\subsection{Energy-based Model (EBM)}
Energy-based models define an explicit density over the observed variables as: 
\vspace{-10pt}
{\fontsize{\mathfnt}{\bse}\selectfont
\begin{align}
    p_\theta (x) = \frac{1}{Z_\theta} e ^ {-f_\theta(x)}
\end{align}}

where $f_\theta : \R^D \rightarrow \R$ is the energy function and the normalization constant $Z_\theta = \int e^{-f_\theta(x)} dx$. The energy function $f_\theta(x)$ can be modeled by any neural network. This makes EBMs a class of very flexible generative models and applicable to a wide variety of applications \cite{ingraham, du_ebm}. 

Although EBMs have many appealing features, the presence of the intractable normalization constant $Z_\theta$ makes training and generation from EBMs hard. For example, the gradient for maximum-likelihood training of EBMs has the following form \cite{DBLP:journals/neco/Hinton02}: 
{\fontsize{9.5}{11.5}\selectfont
\begin{align}
    \nabla_\theta \log p_\theta (x) = -\nabla_\theta f_\theta (x) + \E_{p_\theta(x')} \left[ \nabla_\theta f_\theta(x') \right]
    \label{eq:ebm_grad}
\end{align}
}
The exact gradient estimation needs samples from the model density $p_\theta$ which is intractable in practice. Due to this reason, MCMC sampling is used in popular training algorithms such as contrastive divergence \cite{DBLP:journals/neco/Hinton02}.

\vspace{-5pt}
\section{Energy-based Variational Latent Prior (EVaLP)}
Solving the Prior Hole Problem is equivalent to fixing the gap between VAE's prior and the aggregate posterior. Our goal is to learn a flexible prior using an EBM without sacrificing the efficient training and generation of VAEs. In other words, we want to combine the best of both models as listed in Table \ref{tab:vae_vs_ebm}.   
\subsection{Two Stage Approach}
We take a two-stage approach. In the 1st stage, we obtain the $q_{agg}(z)$ by training a vanilla VAE with a base prior (isotropic Gaussian). In the 2nd stage, we transform the simple base prior to a more flexible distribution to match the aggregate posterior $q_{agg}(z)$. 
For this, we model the prior as an exponentially tilted Gaussian,  
\vspace{-6pt}
{\fontsize{\mathfnt}{\bse}\selectfont
\begin{align}
  p_\psi(z) = \frac{1}{Z_\psi} e^{-f_\psi(z)} \ p_0(z) = \Tilde{p}_\psi(z) \ p_0(z) 
\end{align}}
where $p_0(z) = \mathcal{N}(0,I)$ and $Z_\psi = \int e^{-f_\psi(z)} p_0(z) \ dz$ is the normalizing constant. $\Tilde{p}_\psi(z)$ can be interpreted as an unnormalized reweighting factor for the $\N (0, I)$ base prior of vanilla VAE. 

Re-writing the ELBO with an energy-based prior gives: 
{\fontsize{\mathfnt}{\bse}\selectfont
\begin{align}
    \EL'(x) &:= \E_{q_\phi(z|x)}\left[ \log p_\beta (x|z)\right] - \KL \left(q_\phi(z|x)\ ||\ p_\psi(z)\right) \nonumber \\
    &= \EL(x) - \E_{q_\phi(z|x)}[f_\psi(z)] - \log Z_\psi 
    \label{eq:ext_elbo}
\end{align} 
}
We seek the energy function $f_\psi$ that maximizes this ELBO. The 1st term of eq \ref{eq:ext_elbo} is the ELBO of vanilla VAE which we optimize in the 1st stage. Optimizing the 2nd term is equivalent to learning an EBM to fit $q_{agg}(z)$. However the gradient of this term, $\E_{q_\phi(z|x)}\left[\nabla_\psi \log \Tilde{p}_\psi(z)\right]$ is expensive to compute as it needs MCMC samples from the prior distribution $p_\psi(z)$ itself (see eq \ref{eq:ebm_grad}). One of the key motivations of this work is to avoid MCMC altogether while learning the EBM. To this end, we leverage the variational form of the log-normalizing constant as described next. 
  
\subsection{Learning the Prior using Variational Form of the Log-normalizing Constant}
The presence of the normalizing constant necessitates the requirement of MCMC during maximum likelihood training of the extended-ELBO in eq \ref{eq:ext_elbo}. We can bypass this issue by using the log-normalization constant in the variational form following the steps in  \cite{grathwohl2021no}:
{\fontsize{\mathfnt}{\bse}\selectfont
\begin{align}
    \vspace{-10pt}
    \log Z_\psi &= \max_{p_g} -\KL(p_g || p_\psi) + \log Z_\psi \nonumber \\
    &= \max_{p_g} \ -\E_{p_g(z)}[f_\psi(z)] + \mathcal{H}(p_g) +  \E_{p_g(z)}[\log p_0(z)]  \label{eq:z_var_from} 
    \vspace{-20pt}
\end{align}}
where $p_g$ is a variational distribution induced by an auxiliary sampler $g$ and $\mathcal{H}(p_g)=-\E_{p_g(z)}[\log p_g(z)]$ is the entropy of $p_g$. See Derivations in Appendix for detailed steps. Now let g be parameterized by $\alpha$. 
Using the above form, we can derive an upper bound for the ELBO $\EL'$ as:
{\fontsize{\mathfnt}{\bse}\selectfont
\begin{align}
    \EL'(x) 
    &\leq \EL(x) - \E_{q_\phi(z|x)}[\ f_\psi(z)] + \E_{p_{g_\alpha}(z)}[f_\psi(z)] \nonumber\\ 
    & \quad  + \KL \left( p_{g_\alpha} \ || \ p_0 \right) := \EL^{up}_{EVaLP}(x)
    \label{eq:ub_ext_elbo}
\end{align}}
Note that $\EL(x)$ is a constant w.r.t both $f_\psi$ and $g_\alpha$. $\EL^{up}_{EVaLP}(x)$ can be taken as a new training objective. 
From Eq. \ref{eq:z_var_from} we can see that the inequality above is tight and achieves equality when $p_{g_\alpha}$ = $p_\psi$. The presence of the KL term requires evaluation of $p_{g_\alpha}$ and efficient sampling from it. In this work, we choose $g_\alpha$ to be a normalizing flow model \cite{dinh2017density} which satisfies both the criteria. We minimize the objective w.r.t $g_\alpha$ first and then maximize w.r.t $f_\psi$. So at this stage, our prior learning objective becomes: 
{\fontsize{\mathfnt}{\bse}\selectfont
\begin{align}
    &\max_{p_\psi} \min_{p_{g_\alpha}} \ \EL^{up}_{EVaLP}(x)\label {eq:lebm_obj} 
\end{align}}

\subsection{A More Stable Alternating Optimization Approach}
\label{sec:stability}
Note that, $\EL'(x) \leq \EL^{up}_{EVaLP}(x)$ and $\EL'(x) \leq \log p_\theta(x)$. Maximizaiton of $\EL^{up}_{EVaLP}(x)$ as an ELBO substitute is only valid when it's less or equal to $\log p_\theta(x)$. If the inner minimization in Eq~\ref{eq:lebm_obj} is not performed till optimality, the outer maximization may become unbounded leading to unstable optimization during training. To mitigate this issue, we design a 2nd objective, $\EL^{low}_{EVaLP}(x)$ and we present the following result. 
\begin{proposition}
Let $\EL^{low}_{EVaLP}(x) := \EL^{up}_{EVaLP}(x) - \lambda \E_{\hat{z}\sim \hat{p}(z)}\left[ (||\nabla_{\hat{z}} f_\psi (\hat{z})||_2 - 1)^2 \right]$ where  $\hat{p}$ is implicitly defined by sampling uniformly along straight lines between pairs of points sampled from $q_\phi(z|x)$ and $p_g(z)$. Then for $1$-Lipschitz $f_\psi$ and any $\lambda > 0$,
\begin{enumerate}
    \item Alternatingly optimizing, $\min_{p_{g_\alpha}} \ \EL^{up}_{EVaLP}(x)$ and $\max_{f_\psi} \EL^{low}_{EVaLP}$ is equivalent to optimizing a WGAN with 1-Lipschitz critic $f_\psi$ and decoder $g_\alpha$ with gradient penalty. This WGAN objective has the same solution as Eq~\ref{eq:lebm_obj}.
    \item Let $\alpha^*$ be the optimal $\alpha$ that minimizes Eq~\ref{eq:lebm_obj}, then $\EL^{low}_{EVaLP}(x;\alpha^*) \leq \EL'(x) = \EL^{up}_{EVaLP}(x;\alpha^*)$.
\end{enumerate}
\end{proposition}
The proof of the above proposition directly follows from Eq.~\ref{eq:z_var_from}, \ref{eq:lebm_obj} and the definition of $\EL^{low}_{EVaLP}(x)$ (we also include the proof in the Appendix's Derivations section). 
The first result from Proposition 1 says that our formulation of EBM-based prior is equivalent to training a WGAN in the latent space of the VAE, with an assumption of restricting the EBM class to 1-Lipschitz functions. This allows us to solve the max-min problem in an alternating optimization procedure with two objective functions. The 2nd result shows the advantage of the two objective system.
It shows that in the case of perfect minimization, $\EL^{low}_{EVaLP}$ always remains lower than $\EL'(x)$ for any $\lambda > 0$. But even in the case of imperfect minimization (which is usually the case during training), a large enough $\lambda$ can be chosen such that $\EL^{low}_{EVaLP}(x;\alpha) < \EL'(x)$ throughout the training process and this leads to a more stable optimization. For these reasons, our proposed alternating optimization procedure takes the following form:
{\fontsize{\mathfnt}{\bse}\selectfont
\begin{align*}
    &\text{Step 1:} \quad \alpha^* = \argmin_\alpha \EL_{EVaLP}^{up}(x) \\
    &\text{Step 2:} \quad \max_{\psi} \ \EL_{EVaLP}^{low}(x; \alpha^*)
    \vspace{-10pt}
\end{align*}}
\subsection{Test Time Sampling from the Prior}
 The variational sampler $p_{g_\alpha}$ together with the energy function provides us two ways of sampling from the prior during generation. \\
\textbf{Fast Approximate Sampling:}
Utilizing the fact that $g_\alpha$ is an amortized sampler for the EBM prior $p_\psi$, we have $\{z^m\}_{m=1}^M \sim p_\psi(z)$ if $\Tilde{z}^m \sim \N (0, I)$ and  $z^m = g_\alpha(\Tilde{z}^m), \ m = 1,2,..,M$.
This sampling process is fast as it needs only one forward pass of  $\N (0, I)$ sampled batch through $g_\alpha$. \\
\textbf{Accurate Sampling using Sampling-Importance-Resampling (SIR):} 
At optimality, $\E_{q_{\text{agg}}}[f_\psi(z)] \approx \E_{p_{g_\alpha}}[f_\psi(z)]$ (see Eq.~\ref{eq:lebm_obj}).
Due to inherent nature of neural network training, $p_{g_\alpha}$ may  not be equal to $q_{\text{agg}}$, but $f_\psi$ still has the information of how close (or far) the two distributions are.
We exploit this via energy-guided SIR (SNIS in \cite{eim}) using $p_{g_\alpha}(z)$ as the proposal.
For this, we first sample $\{z^m\}_{m=1}^M$ from $p_{g_\alpha}$. Then we select one of these $M$ samples using the corresponding importance weights $\frac{p_\psi(z^m)}{\sum_{i=1}^M p_\psi(z^m)}$. We use $p_\psi(z^m) = \frac{e^{-f\psi(z^m)} p_{g_\alpha}(z^m)}{\hat{Z}_\psi}$ where we approximate the normalizing constant as $\hat{Z}_\psi = \frac{1}{N} \sum_{i=1}^N e^{-f_\psi(z^l)}, \ \{z^l\}_{l=1}^N \sim p_{g_\alpha}(z)$. 
Note that, unlike methods such as \cite{ncp-vae} that use Gaussian  proposal, we can use a learned variational sampler induced by $p_{g_\alpha}$ as the proposal distribution. 

\subsection{Modeling Hierarchical Priors}
\label{sec:hvae}
In this section we discuss how prior learning using EVaLP can be extended to Hierarchical VAE (HVAE) models. In deep hierarchical VAEs, the expressiveness of prior and posterior is increased by partitioning the latent variables into $L$ disjoint groups $\{z_1, z_2, ..., z_L \}$. 
\begin{table}[ht]
    \centering
    \scriptsize
        \centering
        \begin{tabular}{@{}llll@{}}
            \toprule
             & MNIST & Celeba64 & CIFAR10 \\ \midrule
            VAE & 19.34 & 48.73 & 108.6  \\
            WAE $^\dagger$ & 20.42 & 53.67 & 117.44 \\
            2stage-VAE $^\dagger$ & 18.81 & 49.70 & 109.77 \\
            RAE-L2 (Gauss) $^\dagger$ & 22.22 & 51.13 & 80.80 \\
            RAE-L2 (GMM) $^\dagger$ & 8.69 & 47.97 & 74.16 \\
            NCP-VAE & 13.64 & 41.30 & 92.33  \\
            Latent Flow & 15.05 & 39.90 & 84.21  \\
            \midrule 
            {\prior} (ours) & 9.21 & 38.96 & 77.17 \\
            {\snis} (ours) & 8.10 & 35.90 & 76.43  \\ 
            \bottomrule
        \end{tabular}
        \caption{FID comparison of the proposed method against VAE-based methods with single latent group such as WAE \cite{wae}, 2stage-VAE \cite{2svae}, RAE-L2 \cite{rae}, and NCP-VAE \cite{ncp-vae} on MNIST, CIFAR10 and Celeba64. $\dagger$ implies results are borrowed from \cite{rae}.}
        \vspace{-0.15in}
        \label{tab:main_result_1}
    \end{table}
The prior and the posterior distributions are represented by $p(z) = \prod_l p(z_l | z_{<l})$ and $q(z|x) = \prod_l q(z_l | q_{<l},x)$ where $z_{<l}$ is the collection of latents up to $(l-1)^{th}$ group. The conditionals in these two distributions are usually modeled as factorial Normal distributions.

In generalizing our approach to hierarcical VAE models, we closely follow the generation and inference procedure from \cite{lars}. We apply EVaLP on the top-most latent layer $z_L$ and model the $p(z_L)$ and the joint prior $p(z)$ as: 
\vspace{-10pt}
{\fontsize{\mathfnt}{\bse}\selectfont
\begin{align}
    p(z_L) &=  p_\psi(z_L)=\frac{1}{Z_\psi} e^{-f_\psi(z_L)} \ p_0(z_L) \nonumber \\ 
    p(z) &= p_\psi(z_L) \prod_{l=1}^{L-1} p(z_l | z_{>l}) \label{eq:factored_prior}
\end{align}}
Note that, although this puts EVaLP in the same group of \cite{snis, lars} where only the last latent group of a hierarchical VAE can be modeled, EVaLP maintains faster sampling due to the learned variational sampler. 

\section{Experiments}
\label{sec:expt}

Our experiments aim to answer the following questions: \\
(a) How does learning the {\prior} prior improve VAE's image generation performance? \\ 
(b) To what extent does {\prior} mitigate prior holes? \\ 
(c) How robust is {\prior}, i.e., how does the prior hole problem from the 1st stage affect the 2nd stage performance? \\
(d) How does sampling time improve with {\prior} and {\snis}? 


\textbf{Metrics:} We use the {\em Fréchet Inception Distance} (FID)~\cite{fid} to assess generation quality.
We use {\em Maximum Mean Discrepancy} (MMD) to measure the distance between the prior and posterior, thereby quantifying the prior hole issue.
We use CPU time to compare the latency of different methods.
All experiments were performed in an NVIDIA GeForce RTX 3090 GPU.
All architecture and training details, including the energy function, the RealNVP sampler model, and various hyperparameters are reported in 
Appendix \ref{app:architectures}.

\begin{figure*}[] 
    \centering
        \includegraphics[width=0.8\textwidth]{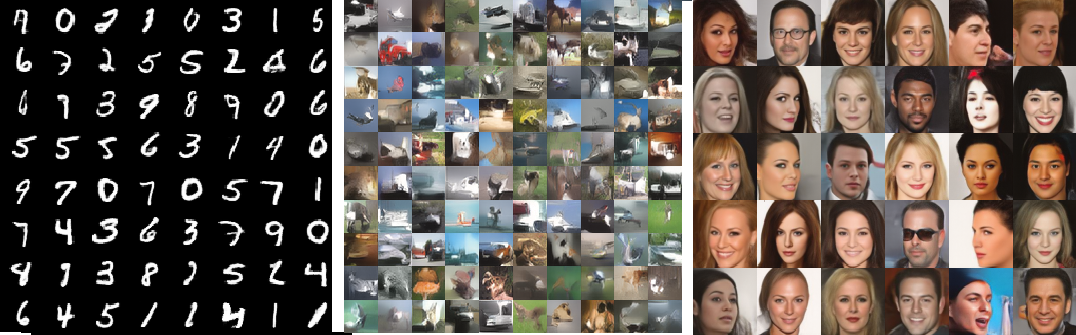}
    \caption{Sampled images from {\snis} prior  trained on MNIST (Table \ref{tab:main_result_1}), CIFAR10 and Celeba64 (Table \ref{tab:main_result_2}).}
    \vspace{-10pt}
    \label{fig:gen_image} 
  \end{figure*}

\subsection{Image Generation}
A VAE model equipped with {\prior} should see inprovements in generation capabilities. To see this, firstly, we compare {\prior} with other small 2-stage prior learning approaches (see. Table \ref{tab:main_result_1}). 
    \begin{table}
        \centering
        \scriptsize
    \begin{tabular}{@{}p{3.5cm}ccp{1.5cm}@{}}
    \toprule
    & \multicolumn{2}{c}{FID} & NFE(FP/BP) \\ 
    \cmidrule(lr){2-3}
    & Celeba64 & CIFAR10 & \\ \midrule
    NVAE $^*$ \cite{nvae} & $15.69$ & $47.10$ & $1/0$ \\
    \midrule 
    \multicolumn{4}{@{}l}{\textbf{Single latent EBM w/o MCMC}} \\
    1level-NCP-VAE & $15.0$ & $43.26$ & $500/0$\\
    {\prior} (ours) & $13.71$ & $42.70$ & $1/0$  \\
    {\snis} (ours) & $13.40$ & $42.30$ & $500/0$ \\ 
    \midrule
    \multicolumn{4}{@{}l}{\textbf{Single latent EBM w/ MCMC}} \\
    LEBM$^\dagger$ \cite{latent-ebm} & 37.87 & 70.15 & 60/60\\
    Adaptive CE $^\dagger$ \cite{adaptive_ce} & 35.38 & 65.01 & 400/400 \\
    Two-Flows \cite{twoflows} & $33.64$ & $66.41$ & 800/800\\
    \midrule 
    \multicolumn{4}{@{}l}{\textbf{Multi-latent EBMs}} \\
    NCP-VAE \cite{ncp-vae} & 5.25 & 24.08 & $150000/0$  \\
    EMLG \cite{emlg} & - & 11.34 & - \\
    EHMLG \cite{ehmlg} & 32.15 & 63.42 & -\\
    \midrule
    \multicolumn{4}{@{}l}{\textbf{Other EBMs on data-space}} \\
    Dual-MCMC \cite{dual_mcmc} & 5.15 & 9.26 & 30/30 \\
    VAEBM \cite{vaeebm} & $5.31$ & $12.19$ & $16/16$\\
    Div-triangle \cite{div_triangle} & 31.92 & - & $1/0$\\
    DDAEBM \cite{ddaebm} & 10.29 & 4.82 & 4/4\\
    \midrule
    \multicolumn{4}{@{}l}{\textbf{GANs \& score-based}} \\
    Style GAN-v2 \cite{stylegan2} & - & 3.26 & 1/0 \\
    NCSN-v2 \cite{ncsn_v2} & 10.23 & 10.87 & 0/1000\\
    DDPM \cite{ddpm} & - & $3.17$ & 0/1000 \\ 
    \bottomrule
\end{tabular}
        \caption{FID score and sampling time NFE (Neural Function Evaluation) comparison on CIFAR10 and Celeba64. For NFE we consider both function evaluation in  Forward pass (FP) and gradient calculation in Backward pass (BP). Model$^*$ indicates backbone for {\prior} and models with $^\dagger$ do not use a multi-layer generator. }
        \label{tab:main_result_2}
     \vspace{-12pt} 
     \end{table}
Secondly, we present results on extending {\prior} to deep hierarchical VAEs (see Table \ref{tab:main_result_2}).

\subsubsection{VAE with single latent group}
In Table \ref{tab:main_result_1} we compare our prior learning method with other baselines based on VAEs with a single latent group. We re-implement the VAE for our experiments in Pytorch \cite{pytorch} by closely following the setup provided in \cite{rae}. Note that, all the baselines in this table share the same VAE architecture. Both NCP-VAE and our {\prior} method are based on EBMs; we use the same energy model architecture for fair comparison. While NCP and {\prior} use the same trained VAE to learn the prior in 2nd stage, for WAE, 2stage-VAE, and RAE-L2, we borrow the reported numbers from \cite{rae}. The baselines all aim to fit a flexible prior, either during training or in a second stage (when the stochastic/deterministic auto-encoder has been trained). As an ablation, we also implement a baseline (Latent Flow) with the same Flow model in the latent space to learn the aggregate posterior distribution.  

Table \ref{tab:main_result_1} reports FID scores for the three datasets. The FID numbers from the implemented models are average of three runs of the FID calculation. 
The image generation quality from {\prior} shows consistent improvement over the baselines. 
With improved sampling, {\snis} (with $500$ proposal samples) brings further gains. 
In MNIST, {\prior} outperforms  most of the baselines (except for RAE-L2(GMM)) while bringing base VAE's FID from $19.21$ to $9.21$. {\snis} pushes FID further to $8.10$, achieving the best FID. 
Similarly, in CelebA, {\prior} improves the base VAE's performance from $48.73$ to $38.96$, outperforming all the baselines, with further improvement with {\snis}.

Results on CIFAR-10 are also encouraging.
{\prior} improves the base VAE's performance from $108.6$ to $77.17$ with {\prior} and to $76.43$ with {\snis}. {\snis}'s performance is slightly worse than RAE-L2 with a GMM prior. 
The likely reason for this is that {\prior} uses a pre-trained decoder from the base VAE, and it's final performance is bottlenecked by the decoder's performance. 

Finally, we observe that {\prior} significantly outperforms the Latent Flow model in all three datasets. This highlights the effectiveness of an energy-based prior over a straight-forward flow-based latent prior.

\subsubsection{Hierarchical VAEs}
We extend {\prior} to hierarchical VAEs (HVAEs) using the procedure described in section \ref{sec:hvae} with NVAE \cite{nvae} as a base VAE model (using available code\footnote{https://github.com/NVlabs/NVAE}). 
Performance of {\prior} against other baselines are evaluated using Celeba64 and CIFAR10 datasets. For Celeba64, we use $15$ latent groups and for CIFAR10, we use $30$ latent groups in our base NVAE model. 

In Table~\ref{tab:main_result_2}, we show FID scores  and sampling time complexity in Neural Function Evaluations(NFE) (in number of Forward and backward evaluations) of different latent EBM prior models as well as EBM on data-space. 
Note that, score function calculation in EBM is equivalent to $1$ FP and $1$ BP. 
We also compare these models against more competitive baselines such as GANs and score-based models.
Note that, although Row 3 models (Single latent EBM w/ MCMC) learn an EBM on a single latent group similar to {\prior}, LEBM~\cite{latent-ebm} and Adaptive-CE~\cite{adaptive_ce} are decoder-based generative models and they do not use an inference model for posterior sampling. 
Hence, they require MCMC during both training and generation and become computationally heavy. 
The (Multi-latent EBMs) models use all the latent-groups of a HVAE model to build the EBM prior. 
The additional performance gain in these models comes at the cost of high computational cost. 
While NCP-VAE~\cite{ncp-vae} trains $30$ binary classifiers to jointly model all the latent-groups, EMLG~\cite{emlg} and EHMLG~\cite{ehmlg} both require expensive MCMC during training and generation. 

{\prior} improves FID of the base NVAE model on both datasets and achieves the best performance among the computationally inexpensive and faster sampling prior learning approaches (see Row 2 in Table~\ref{tab:main_result_2}). 
Note that NCP-VAE \cite{ncp-vae} trains a prior using all latent groups  of the NVAE model. 
For better comparison, we also train an NCP model ($1$level-NCP-VAE) on the last latent layer of the NVAE model with the same smaller energy model architecture used by {\prior}. 
While {\prior} is more robust against prior holes and provides faster sampling (see experiment details in section \ref{sec:robustness} \& \ref{sec:sampling_efficiency}), it also outperforms  NCP-VAE in modeling a single latent group. 
\vspace{-5.5pt}
\subsection{Qualitative results}
\begin{figure}[hbt] 
\vspace{-5pt}
    \centering
        \includegraphics[width=0.48\textwidth]{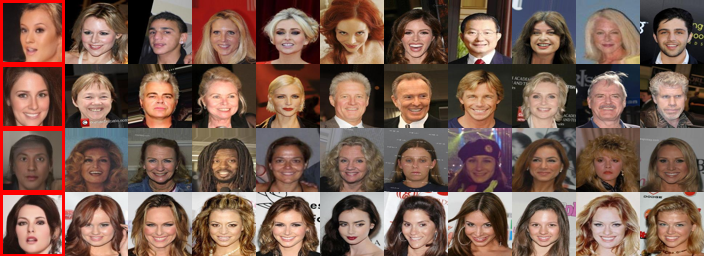}
    \caption{The left column (in red box) corresponds to query images that are generated from the model. The images to the right of the queries are its nearest neighbors from the train set.}
    \label{fig:nn} 
    \vspace{-10pt}
  \end{figure}

Fig.~\ref{fig:gen_image} presents randomly generated image samples from
VAE with {\snis} prior trained on MNIST (\ref{tab:main_result_1}), CIFAR10, and Celeba64 (\ref{tab:main_result_2}) dataset. For the purpose of visualization, for CIFAR10 and Celeba64, the images from base NVAE are generated by scaling down the temperature to $t=0.7$ in the conditional prior distributions following the procedure from \cite{nvae}. More images are included in the appendix \ref{app:visualization} for better assessment of the generation capabilities.\\
\textbf{Nearest neighbors to assess overfitting:}
\label{sec:nn}
To qualitatively assess whether our {\prior} model has over-fitted to the dataset, we show the $10$ nearest neighbor images for each image generated by {\prior} after training on CelebA. 
We use the KD-Tree algorithm to generate the nearest neighbors.
The generated images are compared using Euclidean distance after projecting to a $100$-dimensional PCA plane. All the images from the train split were cropped to $64\times 64$ and scaled to $[0,1]$.
Fig.~\ref{fig:nn} shows the results -- the left most column shows the generated images and the corresponding row lists the $10$ nearest neighbors in the training dataset. We observe that the nearest neighbors are quite different from the query image, suggesting that the learned {\prior} is not prone to overfitting. 

\vspace{-3pt}
\subsection{Robustness  against Prior holes} 
\textbf{Prior Holes after the Second Stage:}
We study the extent to which {\prior} can mitigate the prior hole problem in comparison to the most competitive baseline NCP (which was also aimed at mitigating prior holes).
For this experiment, we use the same settings used in Table \ref{tab:main_result_1}. We use Maximum Mean Discrepancy (MMD) \cite{mmd} to measure this (mis)match.
Table \ref{tab:mmd_res} shows that while NCP improves over VAE, {\prior} and {\snis} improve beyond NCP, achieving the lowest score of $0.2448$. \\   
\begin{table}[h]
\vspace{-10pt}
\centering\
\scriptsize
\begin{tabular}{@{}lcccc@{}}
    \toprule
    Model & VAE & NCP & \prior & {\snis} \\ \midrule
    MMD $(\downarrow)$ & 0.3532 & 0.2978 & 0.2787 & 0.2448 \\ \bottomrule
\end{tabular}
\vspace{-4pt}
\caption{MMD between $q_{agg}(z)$ and $p_0(z)$.}
\vspace{-10pt}
\label{tab:mmd_res}
\end{table}

\begin{figure}[htp]
\vspace{-5pt}
    \centering
        \centering
\includegraphics[width=0.4\textwidth]{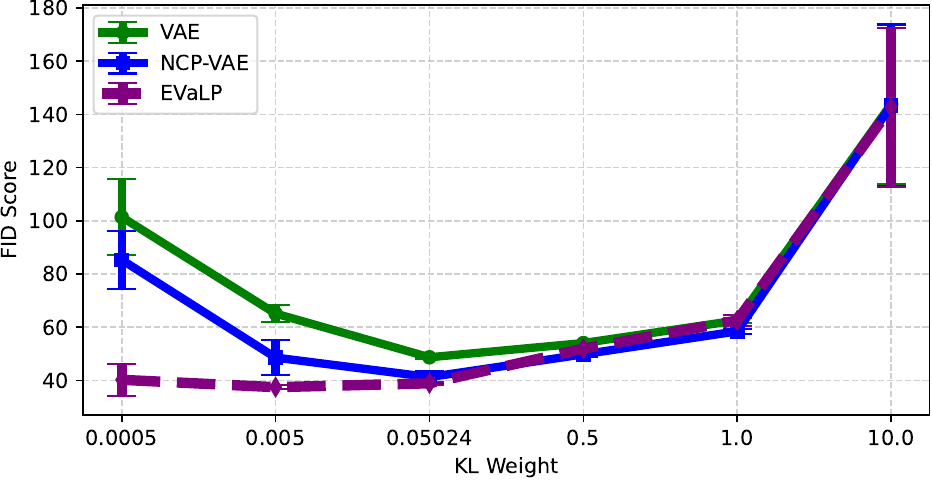}
        \vspace{-0.05in}
        \caption{Comparison of impact of the amount of prior holes in 1st stage on 2nd stage performance.}
        \label{fig:kl_exp}
    \vspace{-13pt}
\end{figure}

\textbf{Robustness Against Severity of Prior Holes:}
\label{sec:robustness}
It is common practice to tune the weight of the KL loss term in VAE's implementation; the weight is a knob to balance between the VAE's reconstruction quality and generative performance. 
Since this KL weight controls how much the aggregate posterior matches the prior, it also essentially controls the severity of prior holes.
We design an experiment where we train multiple VAEs in the 1st stage with different weights for the KL loss. 
Then, in the 2nd stage, we learn {\prior} on each of these VAE models. Fig \ref{fig:kl_exp} compares the FID scores between {\prior} and NCP against increasing KL weights. 

For a very high KL weight, the $q_{agg}$ approximately matches the base prior $p_0$ (i.e., negligible prior hole), hence all three methods attain similar FID.
These FID scores are obviously poor since the VAEs have not optimized well for the reconstruction loss.
However, for low KL weights, {\prior} significantly outperforms both VAE and NCP. 
This also confirms the fact that, the energy-based reweighting factor in NCP doesn't always converge to the true density ratio $\frac{q_{agg}(z)}{p_0(z)}$; the issue is pronounced when the mismatch between the two distributions is high. 
In contrast, Figure \ref{fig:kl_exp} indicates that {\prior} can robustly approximate the aggregate posterior. 

\subsection{Sampling Efficiency from Prior}
\label{sec:sampling_efficiency}
Our method uses a separate sampler network $g_\alpha$ to amortize the sampling process of the EBM prior.  Table \ref{tab:sampling_eff} compares the sampling efficiency of {\prior} and {\snis} with NCP. Note that, NCP uses an energy function-based SIR for sampling from the prior. While sampling from {\prior} is just a forward pass of $\N(0,I)$ sampled latent vectors through the sampler model, {\snis} uses SIR using $p_{g_\alpha}$ as the proposal distribution. We compare the average time required to sample 1 image from VAE trained with different priors on a single AMD 3960X 24-core CPU. For each model's prior, we draw 50,000 samples to calculate the sampling time per sample. For this experiment, we use the models trained on CelebA in Table \ref{tab:main_result_1} and sample a $64\times64$ image with varying numbers of proposals using SIR.  
Comparing Table \ref{tab:sampling_eff} and 
Table~\ref{tab:main_result_1}, we see that
{\prior} offers the best balance by providing good generation quality at the cost of the shortest sampling time. Our optional sampling method {\snis} achieves the best generation at the cost of increase in sampling time compared to NCP.

\begin{table}[h]
\centering
\scriptsize
\setlength{\tabcolsep}{8pt}
\renewcommand{\arraystretch}{0.8}
\begin{tabular}{l|cccccc}
\toprule
\textbf{Method} & 500 & 1000 & 2000 & 3000 & 4000 & 5000 \\
\midrule
NCP & 0.82 & 1.35 & 2.56 & 3.52 & 4.57 & 5.55 \\
EVaLP & 0.79 & - & -& -& -& - \\
EVaLP+SIR & 3.37 & 5.58 & 9.24 & 12.26 & 16.24 & 19.36 \\
\bottomrule
\end{tabular}
\vspace{-4pt}
\caption{Seconds per sample (in \( \times 10^{-3} \)) for different methods and numbers of proposals samples. Note, EVaLP doesn't require any proposal to sample from it.}
\label{tab:sampling_eff}
\vspace{-10pt}
\end{table}

\vspace{-5pt}
\section{Conclusion}
The prior hole problem is one of the key weaknesses of a VAE. We attempt to mitigate this problem by learning an energy-based {\em flexible} prior. We leverage the variational form of the EBM's log-normalizing constant to introduce a sampler network, eliminating the need for MCMC sampling. Our method is similar to training a WGAN in the latent space of a VAE. Upon convergence, our method provides two ways of sampling from the prior: (a) Fast sampling using the sampler network and (b) Accurate sampling using SIR where the sampler offers the proposal distribution. Finally, we demonstrate that while our learned prior {\prior} attains competitive performance in Hierarchical VAEs in the group of fast-sampling and less compute intensive priors, it 
 attains superior or competitive generation performance over other two-stage prior learning methods in single latent group VAE models.
 
{
    \small
    \bibliography{main}
}
\clearpage
\newpage
\appendix
\vspace{1em}

\section{Appendix}
\label{sec:append}

\subsection{Derivations}
\label{app:derivations}
In this section, we give more details on some equations used in the main text of the paper for completeness and include the proof for Proposition 1. 
\subsubsection{Maximize ELBO w.r.t Prior:}
Let, $q_{agg}(z) = \E_{p_d}(x) [q_\phi(z|x)]$. Then following the steps in \cite{ncp-vae} optimization of VAE's average ELBO, $\E_{p_d(x)}[\EL(x)]$ w.r.t prior $p_\psi(z)$ is, 
{\fontsize{\mathfnt}{\bse}\selectfont
\begin{align}
    & \argmax_{p_\psi(z)} \E_{p_d(x)} [\EL(x)] \nonumber \\
    &= \argmax_{p_\psi(z)} \E_{p_d(x)} \big[ \E_{q_\phi(z|x)} \left[ \log p_\beta(x|z) \right] - \nonumber \\
     & \quad \quad \quad \KL \left(q_\phi(z|x) \ || \ p_\psi (z)\right) \big] \nonumber \\
     &= \argmin_{p_\psi(z)} \E_{p_d(x)} \left[ \KL \left(q_\phi(z|x) \ || \ p_\psi (z)\right) \right] \nonumber \\
     &= \argmin_{p_\psi(z)} \ \E_{p_d(x)} \left[ -\mathcal{H}(q_\phi(z|x))\right] \nonumber \\
     & \qquad- \E_{p_d(x)} \left[\E_{q_\phi(z|x)}[\log p_\psi(z)] \right] \nonumber \\
     &= \argmin_{p_\psi(z)} -\mathcal{H}(q_{agg}) - \E_{q_{agg}(z)}[\log p_\psi(z)] \nonumber \\
     &= \argmin_{p_\psi(z)} \KL(q_{agg} \ || \ p_\psi)
\end{align}}
Here, the 2nd equality holds since the 1st term doesn't depend on $p_\psi$. Using the same logic, we replace $\E_{p_d(x)} \left[ -\mathcal{H}(q_\phi(z|x))\right]$ with  $-\mathcal{H}(q_{agg}(z))$ in the 4th equality. 
\subsubsection{Steps for deriving Eq~\ref{eq:ebm_grad} (Maximum Likelihood Training of EBM):}
We include the derivation of Eq~\ref{eq:ebm_grad} following the steps in \cite{con_divergence} for completeness. Let $ p_\theta (x) = \frac{1}{Z_\theta} e ^ {-f_\theta(x)}$
where the normalizing constant $Z_\theta = \int e ^ {-f_\theta(x)} dx$. Then, 
{\fontsize{\mathfnt}{\bse}\selectfont
\begin{align}
    \nabla_\theta \log Z_\theta &= \frac{1}{Z_\theta} \nabla_\theta Z_\theta \nonumber \\
    &= \frac{1}{Z_\theta} \nabla_\theta \int e ^ {-f_\theta(x)} \ dx \nonumber \\
    &= \int \frac{1}{Z_\theta} e ^ {-f_\theta(x)} (-\nabla_\theta f_\theta(x)) \ dx \nonumber \\
    &= - \E_{p_\theta(x)}\left[ \nabla_\theta f_\theta (x)\right] \label{app:z_grad}
\end{align}}
Using the gradient of the normalizing constant from eq \ref{app:z_grad} in log-likelihood gradient of $p_\theta(x)$, 
{\fontsize{\mathfnt}{\bse}\selectfont
\begin{align}
    \nabla_\theta \log p_\theta(x) &= \nabla_\theta \left( -f_\theta(x) - \log Z_\theta \right) \nonumber \\
    &= - \nabla_\theta f_\theta(x) + \E_{p_\theta(x')} \left[ \nabla_\theta f_\theta (x') \right]
\end{align}}

\subsubsection{Derivation of Eq~\ref{eq:z_var_from} (Variational form of the log-nomalizing constant):} Here we show the detailed steps for deriving Eq.~\ref{eq:z_var_from} in the main paper. Using the fact that $KL(p \ || \ q) \geq 0$ for any distributions $p$ and $q\geq 0$, the term $\log z_\psi$ can be written as:
{\fontsize{\mathfnt}{\bse}\selectfont
\begin{align}
    \log Z_\psi &= \max_{p_g} -\KL(p_g || p_\psi) + \log Z_\psi \nonumber \\
    &= \max_{p_g} \int p_g(z) \log \left(\frac{e^{-f_\psi(z)}p_0(z)/Z_\psi}{p_g(z)} \right) \ dz \nonumber \\
    & \quad \quad \quad + \log Z_\psi \nonumber \\
    &= \max_{p_g} -\int p_g(z) f_\psi(z)\ dz - \int p_g(z) \log p_g(z)\ dz \nonumber\\
    & \quad \quad + \int p_g(z) \log p_0(z)\ dz \nonumber \\
    &= \max_{p_g} \ -\E_{p_g(z)}[f_\psi(z)] + \mathcal{H}(p_g) +  \E_{p_g(z)}[\log p_0(z)]
     \label{app:z_var_from} 
\end{align}}
\subsubsection{Steps for Eq~\ref{eq:ub_ext_elbo} in main paper (Upperbound for the extended-ELBO):}
We model {\prior} prior as an exponentially tilted Gaussian,  
\vspace{-6pt}
{\fontsize{\mathfnt}{\bse}\selectfont
\begin{align}
  p_\psi(z) = \frac{1}{Z_\psi} e^{-f_\psi(z)} \ p_0(z) = \Tilde{p}_\psi(z) \ p_0(z) 
\end{align}}
where $p_0(z) = \mathcal{N}(0,I)$ and $Z_\psi = \int e^{-f_\psi(z)} p_0(z) \ dz$ is the normalizing constant. 
The ELBO term for a vanilla-VAE (with $p_0 = \N(0, I)$ as the prior) is given by:
{\fontsize{8}{12}\selectfont
\begin{align}
    \EL(x) = \E_{q_\phi(z|x)} \big[\log p_\beta(x|z)\big] - 
    \KL\big(q_\phi(z|x) \| p_0(z)\big) 
\end{align}
}
Re-writing this ELBO with the energy-based prior defined above gives:
{\fontsize{\mathfnt}{\bse}\selectfont
\begin{align}
    \EL'(x) &= \EL(x) + \E_{q_\phi(z|x)}\left[\ \log\Tilde{p}_\psi (z)\ \right] \nonumber \\
    &= \EL(x) - \E_{q_\phi(z|x)}[f_\psi(z)] - \log Z_\psi \nonumber \\
    &\underset{(i)}{=} \EL(x) - \E_{q_\phi(z|x)}[f_\psi(z)] - \nonumber \\
    & \left[\max_{p_{g_\alpha}} \ -\E_{p_{g_\alpha}(z)}[f_\psi(z)] + \mathcal{H}(p_{g_\alpha}) +  \E_{p_{g_\alpha}(z)}[\log p_0(z)] \right] \nonumber \\
    &\leq \EL(x) - \E_{q_\phi(z|x)}[\ f_\psi(z)] + \E_{p_{g_\alpha}}[\ f_\psi(z)] + \E_{p_{g_\alpha}}[\log p_{g_\alpha}] \nonumber \\
    & \qquad -\E_{p_{g_\alpha}}[\log p_0] \nonumber \\
    &= \EL(x) - \E_{q_\phi(z|x)}[\ f_\psi(z)] + \E_{p_{g_\alpha}(z)}[f_\psi(z)] \nonumber\\ 
    & \quad  + \KL \left( p_{g_\alpha} \ || \ p_0 \right)
    \label{app:ub_ext_elbo}
\end{align}}
where in $(i)$ we used the variational form of the log-normalizing constant from Eq.~\ref{app:z_var_from}. 
\vspace{5pt}

\subsubsection{Proof of proposition 1:}
We restate the Proposition 1 from the main paper and include the proof here. 
\textbf{Proposition:} 
Let $\EL^{low}_{EVaLP}(x) := \EL^{up}_{EVaLP}(x) - \lambda \E_{\hat{z}\sim \hat{p}(z)}\left[ (||\nabla_{\hat{z}} f_\psi (\hat{z})||_2 - 1)^2 \right]$ where  $\hat{p}$ is implicitly defined by sampling uniformly along straight lines between pairs of points sampled from $q_\phi(z|x)$ and $p_g(z)$. Then for $1$-Lipschitz $f_\psi$ and any $\lambda > 0$,
\begin{enumerate}
    \item Alternatingly optimizing, $\min_{p_{g_\alpha}} \ \EL^{up}_{EVaLP}(x)$ and $\max_{f_\psi} \EL^{low}_{EVaLP}$ is equivalent to optimizing a (KL regularized) WGAN with 1-Lipschitz critic $f_\psi$ and decoder $g_\alpha$ with gradient penalty. This WGAN objective has the same solution as Eq~\ref{eq:lebm_obj}.
    \item Let $\alpha^*$ be the optimal $\alpha$ that minimizes Eq~\ref{eq:lebm_obj}, then $\EL^{low}_{EVaLP}(x;\alpha^*) \leq \EL'(x) = \EL^{up}_{EVaLP}(x;\alpha^*)$.
\end{enumerate}
\begin{proof} 1) 
    The value function of a WGAN with gradient penalty whose critic is $D(z)$ and generator is $G(z)$ that tries to approximate the real data distribution $P_r(z)$, is given by,
    {\fontsize{\mathfnt}{\bse}\selectfont
    \begin{align}
        L &= \E_{\Tilde{z}\sim P_G}[D(\Tilde{z})] - \E_{z\sim P_r}[D(z)] \ +  \nonumber \\
        & \quad \lambda \E_{\hat{z}\sim P_{\hat{z}}}[(|| \nabla_{\hat{z}} D(\hat{z})||_2 -1 )^2] \label{app:wgan_value_fn}
    \end{align}}
    where it solves $\min_{D} \max_{G} L$ and  $P_{\hat{z}}$ is defined by sampling uniformly along straight lines between pairs of points sampled from the data distribution $P_r$ and the generator distribution $P_G$. On successful training, i.e. when $L\rightarrow 0$, $P_G \approx P_r$ and $D$ is $1$-Lipschitz \cite{improved-wgan}. 

    Now since,  $\EL^{low}_{EVaLP}(x) = \EL^{up}_{EVaLP}(x) - \lambda \E_{\hat{z}\sim \hat{p}(z)}\left[ (||\nabla_{\hat{z}} f_\psi (\hat{z})||_2 - 1)^2 \right]$, the alternating steps of $\min_{g_\alpha} \EL^{up}_{EVaLP}(x)$ and $\max_{f_\psi} \EL^{low}_{EVaLP}(x)$ solves the max-min problem with value function,
    {\fontsize{\mathfnt}{\bse}\selectfont
    \begin{align}
        L' &= \EL^{low}_{EVaLP}(x) \nonumber \\
        &= \EL(x) - \E_{q_\phi(z|x)}[\ f_\psi(z)] + \E_{p_{g_\alpha}(z)}[f_\psi(z)] \nonumber \\ 
        & \quad + \KL \left( p_{g_\alpha} \ || \ p_0 \right) - \lambda \E_{\hat{z}\sim P_{\hat{z}}}[(|| \nabla_{\hat{z}} D(\hat{z})||_2 -1 )^2] \nonumber \\
        & = - \big(\E_{p_{g_\alpha}(z)}[-f_\psi(z)] - \E_{q_\phi(z|x)}[\ -f_\psi(z)] \nonumber \\ 
        & \quad - \KL \left( p_{g_\alpha} \ || \ p_0 \right) + \lambda \E_{\hat{z}\sim P_{\hat{z}}}[(|| \nabla_{\hat{z}} D(\hat{z})||_2 -1 )^2] \big) \nonumber \\ 
        & \qquad + \EL(x) \nonumber \\
        &:= - L'' + \text{ const }
        \vspace{-5pt}
    \end{align}}
    where $\EL(x)$ is a constant w.r.t $f_\psi$ and $g_\alpha$ and we define $L''= \E_{p_{g_\alpha}(z)}[-f_\psi(z)] - \E_{q_\phi(z|x)}[\ -f_\psi(z)] 
         - \KL \left( p_{g_\alpha} \ || \ p_0 \right) + \lambda \E_{\hat{z}\sim P_{\hat{z}}}[(|| \nabla_{\hat{z}} D(\hat{z})||_2 -1 )^2] $. Then the problem is equivalent to  $\max_{f_\psi} \min_{g_\alpha} - L'' + \text{ const } = \min_{f_\psi} \max_{g_\alpha}  L''+\text{const}$. Comparing this with Eq.~\ref{app:wgan_value_fn}, we can conclude that the alternating optmization solves an equivalent (KL regularized) WGAN with gradient penalty where the critic  and the generator are given by $-f_\psi$ and $g_\alpha$ respectively.

    The 2nd part of the proof can be realized by invoking the result that if $f_\psi$ is $1$-Lipschitz and  also the solution of WGAN defined in Eq~\ref{app:wgan_value_fn}, then the gradient penalty term $\E_{\hat{z}\sim P_{\hat{z}}}[(|| \nabla_{\hat{z}} D(\hat{z})||_2 -1 )^2] = 0$ almost everywhere (see Proposition 1 of \cite{improved-wgan}). So, since $\EL^{low}_{EVaLP}(x)=\EL^{up}_{EVaLP}(x)-\lambda \E_{\hat{z}\sim P_{\hat{z}}}[(|| \nabla_{\hat{z}} D(\hat{z})||_2 -1 )^2]$ we have
    {\fontsize{\mathfnt}{\bse}\selectfont
    \begin{align}
        &\max_{f_\psi} \min_{g_\alpha} \EL^{low}_{EVaLP} \nonumber \\
        &= \max_{\substack{f_\psi \\ f_\psi \text{ is 1-Lipschitz}}} \min_{g_\alpha} \EL^{up}_{EVaLP}
    \end{align}}
    i.e. solving the WGAN problem is equivalent to solving Eq~\ref{eq:lebm_obj} by restricting the EBM function class to $1$-Lipschitz functions.
    
    2) Variational form of the log normalizing constant is given by, $\log Z_\psi = \max_{p_g} \ -\E_{p_g(z)}[f_\psi(z)] + \mathcal{H}(p_g) +  \E_{p_g(z)}[\log p_0(z)]$. Also, $\EL_{EVaLP}^{up}(x;\alpha)\geq \EL'(x) \ \forall \alpha$. Now, 
    {\fontsize{\mathfnt}{\bse}\selectfont
    \begin{align*}
        \min_{g_\alpha}& \ \EL_{EVaLP}^{up}(x;\alpha) \nonumber \\
        &=\min_{g_\alpha} \EL(x) - \E_{q_\phi(z|x)}[\ f_\psi(z)] + \E_{p_{g_\alpha}(z)}[f_\psi(z)] \nonumber\\ 
    & \quad  + \KL \left( p_{g_\alpha} \ || \ p_0 \right) \nonumber \\
    &=  \EL(x) - \E_{q_\phi(z|x)}[f_\psi(z)] - \nonumber \\
    & \left[\max_{p_{g_\alpha}} \ -\E_{p_{g_\alpha}(z)}[f_\psi(z)] + \mathcal{H}(p_{g_\alpha}) +  \E_{p_{g_\alpha}(z)}[\log p_0(z)] \right] \nonumber \\ 
    &= \EL(x) - \E_{q_\phi(z|x)}[f_\psi(z)] - \log Z_\psi \nonumber\\
    &= \EL'(x)
    \end{align*}}
    where in the 2nd equality we use the variational form of the log-normalizing constant mentioned above. If this minimum is attained at $\alpha=\alpha^*$, then $\EL_{EVaLP}^{up}(x;\alpha^*) = \EL'(x)$. Again, $\EL_{EVaLP}^{up}(x;\alpha)\geq \EL_{EVaLP}^{low}(x;\alpha) \ \forall \alpha$ and $\forall \ \lambda \geq 0$. Hence, $\EL_{EVaLP}^{low}(x;\alpha^*) \leq \EL'(x) = \EL_{EVaLP}^{up}(x;\alpha^*) $. 
    
\end{proof}

\subsubsection{The Stability Issue of the Upper Bound}
\label{app:stability_issue}
\begin{figure} 
        \centering
        \includegraphics[width=0.45\linewidth]{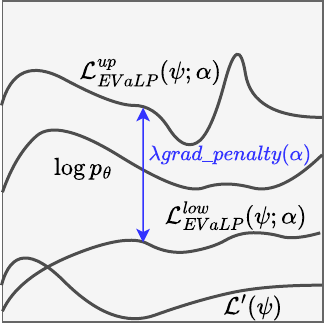} 
        \caption{A scenario where $ \mathcal{L}’_{EVaLP}(\psi; \alpha)$ lies above $\log p_\theta$ even after its inner minimization.}
        \label{fig:stability}
        \vspace{-7pt}
    \end{figure}
We discuss here the stability issue of the max-min objective described in \textit{A More Stable Alternating Optimization Approach} in the main text. Optimization of any ELBO is stable if it remains less than or equal to $\log p_\theta$. Also note that, $\EL'(\psi) \leq \log p_\theta$ and $\EL'(\psi) \leq \EL^{up}_{EVaLP}(\psi, \alpha)$. When perfect  minimization happens in Eq~\ref{eq:lebm_obj}, then  $\EL'(\psi) = \EL^{up}_{EVaLP}(\psi, \alpha^*) \leq \log p_\theta$ (see Proposition 1) and the outer maximization remains stable. But in the case of imperfect inner minimization (which is usually the case in GAN/WGAN training), there could be a scenario where $\EL^{up}_{EVaLP}(\psi, \alpha)$ remains above $\log p_\theta$ (see Figure~\ref{fig:stability}). In that case, the outer maximization of Eq~\ref{eq:lebm_obj} will be invalid and  could potentially become unbounded and unstable. 

Our main idea is that, when a large enough $\lambda$ is used with a negative gradient penalty, we can always keep $\mathcal{L}^{low}_{EVaLP}(\psi, \alpha)$ less than $\log p_\theta$. 
 
In the case of imperfect minimization described above, a large enough $\lambda$ can still keep $\EL^{low}_{EVaLP}(\psi, \alpha)$ lower than $\log p_\theta$. 
In our experiments, we observe training stability with $\lambda=10$ across all the datasets.
\vspace{5pt}  
\subsection{Connection with NCP-VAE\cite{ncp-vae}:} 
Like FCE, NCP-VAE is also based on NCE principle and is situated in the latent space of VAE where the Gaussian prior works as the fixed noise distribution. Because of this fixed noise, NCP-VAE suffers from severity of prior hole problem (demonstrated in Figure 5). This is the reason that it performs poorly compared to EVaLP in Table~\ref{tab:main_result_1} (small VAE model) and  in Table~\ref{tab:main_result_2} (see 1level-NCP-VAE). Although it can be extended to all the latent groups of a HVAE model (such as NVAE) to increase its performance (see NCP-VAE in Table~\ref{tab:main_result_2}), it becomes extremely training and inference heavy (use of 30 classifiers in CIFAR10 and CelebA in Table~\ref{tab:main_result_2}).

\subsection{Network Architectures and Implementation Details}
\label{app:architectures}
\subsubsection{VAE with Single Latent Group}
This section provides the network architectures and training details of {\prior}. We replicate the experiment setup provided in \cite{rae} for the base VAE implementations.
For fair comparison, we use the same energy-function (EBM) architecture in NCP and {\prior}. We give the details of the EBM's energy function, $f_\psi$ and the sampler model $g_\alpha$ below. 

\textbf{Energy Function ($f_\psi$):} 
Table \ref{tab:ebm} provides the energy-function architecture for all three datasets, i.e. CelebA64, CIFAR10 and MNIST. For all the datasets we use an MLP network with $2$ hidden layers. We use $nz=64$; $nd=200$ for CelebA and $nz=128$ and $nd=200$ for CIFAR and $nz=16$ and $nd=128$ for MNIST.

\begin{table}[h]
    \centering
    \begin{tabular}{cc}
        \toprule
        Layer & Output Size \\
        \midrule
         Input & $nz$ \\
         Linear, LReLU & $nd$ \\
         Linear, LReLU & $nd$ \\
         Linear & $1$ \\
         \bottomrule
    \end{tabular}
    \caption{The EBM energy-function architecture for {\prior}. We indicate Leaky-ReLU activation by LReLU with leak-factor 0.01.}
    \vspace{-10pt}
    \label{tab:ebm}
\end{table}

\begin{figure}[h]
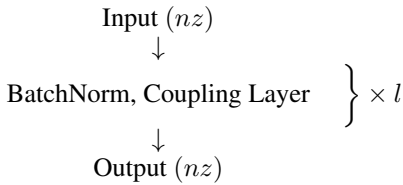

    \centering
    \begin{tabular}{ccc}
    & Input $(nz)$ &  \\
        & $\downarrow$&  \\
         &BatchNorm, Coupling Layer  & $\bigg \}\times l$ \\
         & $\downarrow$&  \\
         & Output $(nz)$ &  \\
    \end{tabular}
    \caption{The architecture of the RealNVP sampler $g_\alpha$ used in experiments of 
 {\prior} on a VAE with a single latent group. A Batch normalization layer followed by a Coupling layer constitutes a Coupling Block. Here, $nz$ is the size of the latent vector and $l$ is the number of Coupling Blocks.}
    \label{fig:flow_arch}
\end{figure}

\textbf{Sampler Function ($g_\alpha$):} 
For {\prior}, the sampler function is implemented with a RealNVP normalizing Flow model \cite{dinh2017density}. Our RealNVP model is a cascade of $l$ coupling Blocks. Each coupling Block is comprised of a Batch Norm layer followd by a Coupling Layer. The Scale network and Translation network of the coupling Layer are implemented as described in \cite{dinh2017density}. A visual representation of our RealNVP sampler network is given in Figure \ref{fig:flow_arch} and the architecture details of Scale and Translation network are provided in Table \ref{tab:coupling}.  
In our experiments, we use  $nz=nf=64, nh=256$ and $l=3$ for CelebA and $nz=nf=128, nh=512$ and $l=4$ for CIFAR dataset. For MNIST a very small latent space is used, where $nz=nf=16$, $nh=128$ and $l=3$.\\

\begin{table}[h]
\begin{tabular}{@{}cc|cc@{}}
\toprule
\multicolumn{2}{c|}{Translation Net} & \multicolumn{2}{c}{Scale Net} \\ \midrule
Layer            & Output Size     & Layer          & Output Size  \\
\midrule
Input            & $nz$              & Input          & $nz$           \\
Linear, Tanh     & $nh$             & Linear, ReLU   & $nh$          \\ 
Linear, Tanh     & $nh$             & Linear ReLU    & $nh$          \\
Linear           & $nf$              & Linear         & $nf$   \\
\bottomrule
\end{tabular}
\caption{The MLP architecture of Translation and Scale network used in Coupling Layer.}
\label{tab:coupling}
\end{table}

\textbf{Optimization:}
For all experiments of NCP and {\prior} we use Adam optimizer. For {\prior}, for both $f_\psi$ and $g_\alpha$, we use learning rate of $2\times 10^{-4}$ for CelebA and $1\times10^{-3}$ for CIFAR. For MNIST we use smaller learning rate; $5\times10^{-4}$ for both energy function and sampler. Gradient penalty weight $\lambda=10$ was used for all the datasets. NCP was trained with learning rate of $1\times 10^{-3}$. For all the experiments, minibatch of size 100 is used. 

For the optimization of energy function and the sampler through max-min game, we update the energy function/critic $f_\psi$ $5$ times for every one update of sampler $g_\alpha$. We found this setting to give consistent and better results in our experiments. In all the datasets, {\prior} is trained for $150$ epochs.

\subsubsection{Hierarchical VAE}
\label{app:arch_hvae}
\begin{figure}
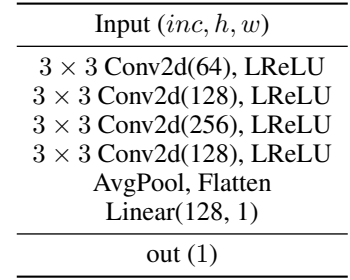

    \centering
    \begin{tabular}{cc}
        \toprule
         Input  ($inc, h, w$) \\
         \midrule
         $3\times3$ Conv2d(64), LReLU \\
         $3\times3$ Conv2d(128), LReLU \\
         $3\times3$ Conv2d(256), LReLU \\
         $3\times3$ Conv2d(128), LReLU \\
         AvgPool, Flatten \\
         Linear(128, 1) \\
         \midrule
         out ($1$)\\
         \bottomrule
    \end{tabular}
    \caption{The energy-function architecture for {\prior} in HVAE experiments. We indicate Leaky-ReLU activation by LReLU with leak-factor 0.01}. 
    \vspace{-15pt}
    \label{fig:ebm_hvae}
\end{figure}

\begin{figure}[h]
    \centering
    \begin{tabular}{ccc}
    & Input $(inc, h, w)$ &  \\
        & $\downarrow$&  \\
         &Coupling Layer ($inc$, $midc$)  & $\big \}\times l_1$ \\
         & $\downarrow$&  \\
         &Coupling Layer ($4\times inc$, $2\times midc$)  & $\big \}\times l_2$ \\
         & $\downarrow$&  \\
         &Coupling Layer ($2 \times inc$, $2\times midc$)  & $\big \}\times l_3$ \\
         & $\downarrow$&  \\
         &Coupling Layer ($2\times inc$,  $2\times midc$)  & $\big \}\times l_4$ \\
         & $\downarrow$&  \\
         & Output $(inc, h, w)$ &  \\
    \end{tabular}
    \caption{The architecture of the RealNVP sampler $g_\alpha$ in an HVAE setup. Here, $l_1, l_2, l_3$ and $l_4$ indicate the number of times the Coupling Layer is repeated.}
    \label{fig:flow_arch_hvae}
\end{figure}

\begin{figure}[h] 
    \centering
        \includegraphics[width=0.2\textwidth]{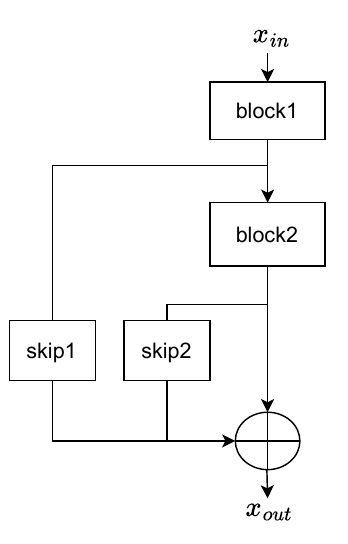}
    \caption{The architecture of a Resnet block used in the coupling layer of RealNVP sampler. See Appendix \ref{app:arch_hvae} for more details on these operations. } 
    \label{fig:resnet_block} 
    \vspace{-6pt}
  \end{figure}
We chose NVAE \cite{nvae} as our base hierarchical VAE model. We train NVAE followed by {\prior} closely following the training details from \cite{ncp-vae}. Our NVAE model trained on Celeba64 uses $15$ latent groups while the CIFAR10 model uses $30$. We give details of the used energy function and sampler function architecture below. \\
\textbf{Energy function}:
We use the same energy function architecture (see Figure \ref{fig:ebm_hvae}) in both Celeba64 and CIFAR10. Each of the Conv2d layer has kernel size $3\times3$, stride $1$ and padding $1$. The number of input channels $inc$ is $20$ for both the datasets.\\
\textbf{Sampler function}: We use a RealNVP model as a sampler function; the architecture is illustrated in Figure \ref{fig:flow_arch_hvae}. The scale and translate network inside the coupling layers are modeled by a Residual network layer shown in Figure \ref{fig:resnet_block}. In a Coupling Layer($inc_i, outc_i$), each block in Figure \ref{fig:resnet_block}, is a cascade of two $3\times3$ Conv2d($outc_i$) layers, where each skip connection block is modeled by a $1\times1$ Conv2d($inc_i$) layer. For Celeba64 we use $l_1=1, l_2=3, l_3=1, l_4=1$ and for CIFAR10, $l_1=3, l_2=3, l_3=3, l_4=1$ is used. \\
Models on both the datasets are trained with Adam optimizer. Learning rate for energy function is $3\times10^{-4}$ and for sampler function we use a smaller rate of $5\times 10^{-5}$. {\prior} was trained for $200$ epochs on both the datasets. 

\begin{figure}[h] 
    \centering
        \includegraphics[width=0.33\textwidth]{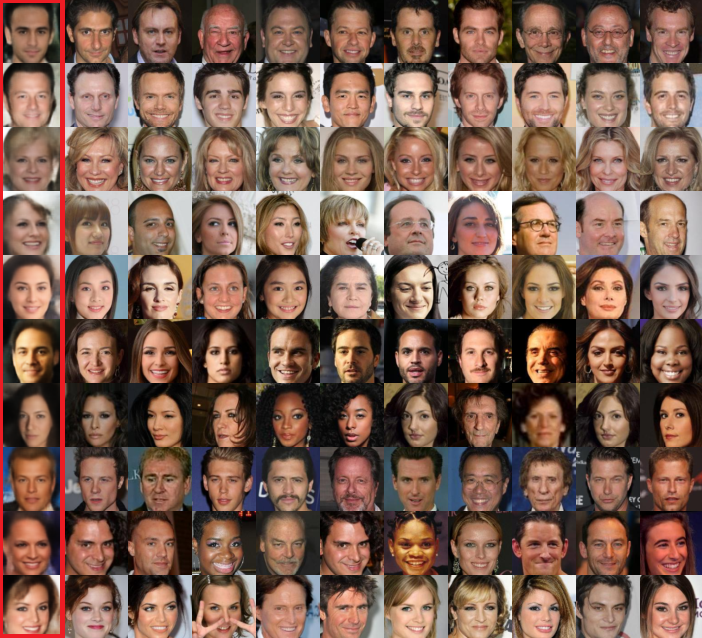}
    \caption{The left-most column (in red box) corresponds to query images that are generated from {\prior} trained on Celeba64 with single latent group (see Table \ref{tab:main_result_1} for the quantitative results). The images to the right of the query are its nearest neighbors from the train set.}
    \label{fig:nn_single_latent} 
    \vspace{-5pt}
  \end{figure}

\begin{figure}[h] 
    \centering
        \includegraphics[width=0.33\textwidth]{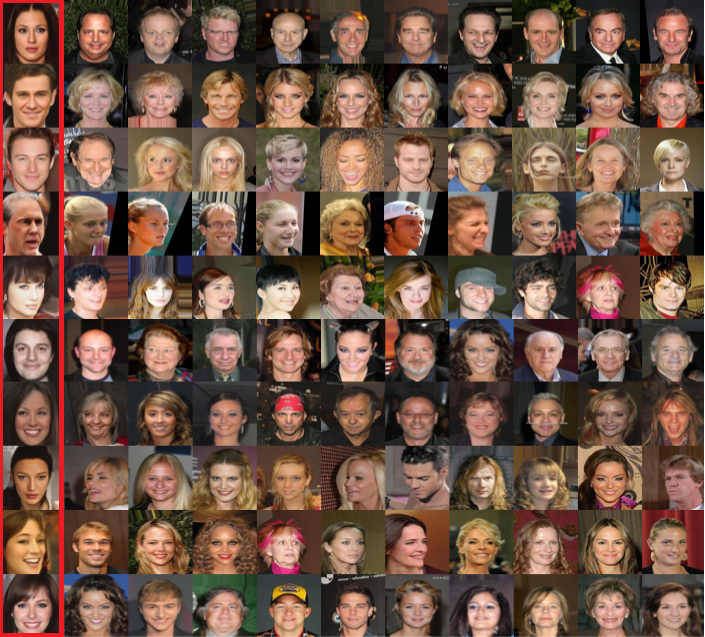}
    \caption{The left-most column (in red box) corresponds to query images that are generated from {\prior} trained on Celeba64 with NVAE as a base in a Hierarchical VAE setting (see Table \ref{tab:main_result_2} for the quantitative results). The images to the right of the query are its nearest neighbors from the train set.}
    \vspace{0pt}
    \label{fig:nn_hvae} 
  \end{figure}

\subsection{More Visualizations}
\label{app:visualization}
\vspace{3pt}
\begin{figure*}[hbt] 
    \centering
        \includegraphics[width=\textwidth]{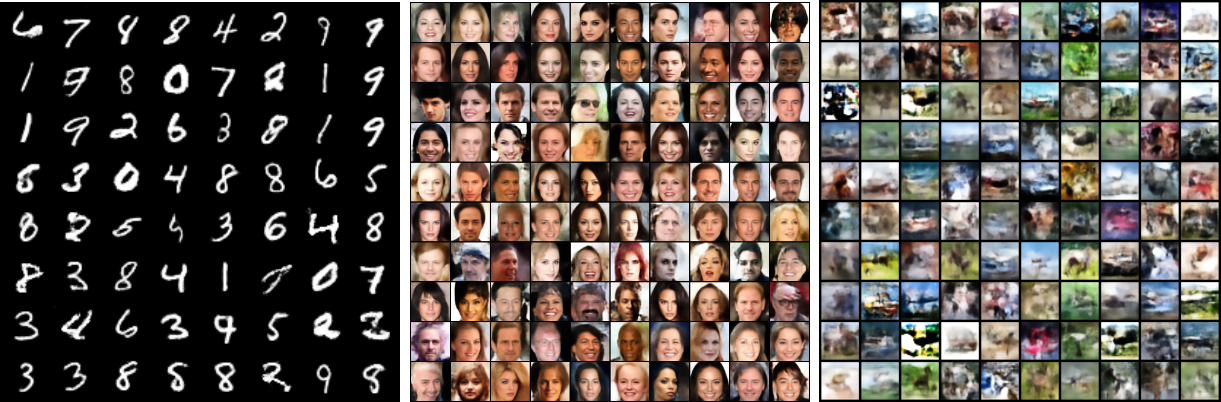}
    \caption{Random generated images from {\prior} trained on MNIST, Celeba64 and CIFAR10 (from left to right) with single latent VAE model (see Table \ref{tab:main_result_1} for quantitative results).}
    \label{fig:gen_small_app} 
  \end{figure*}
  
\begin{figure*}[hbt] 
    \centering
        \includegraphics[width=\textwidth]{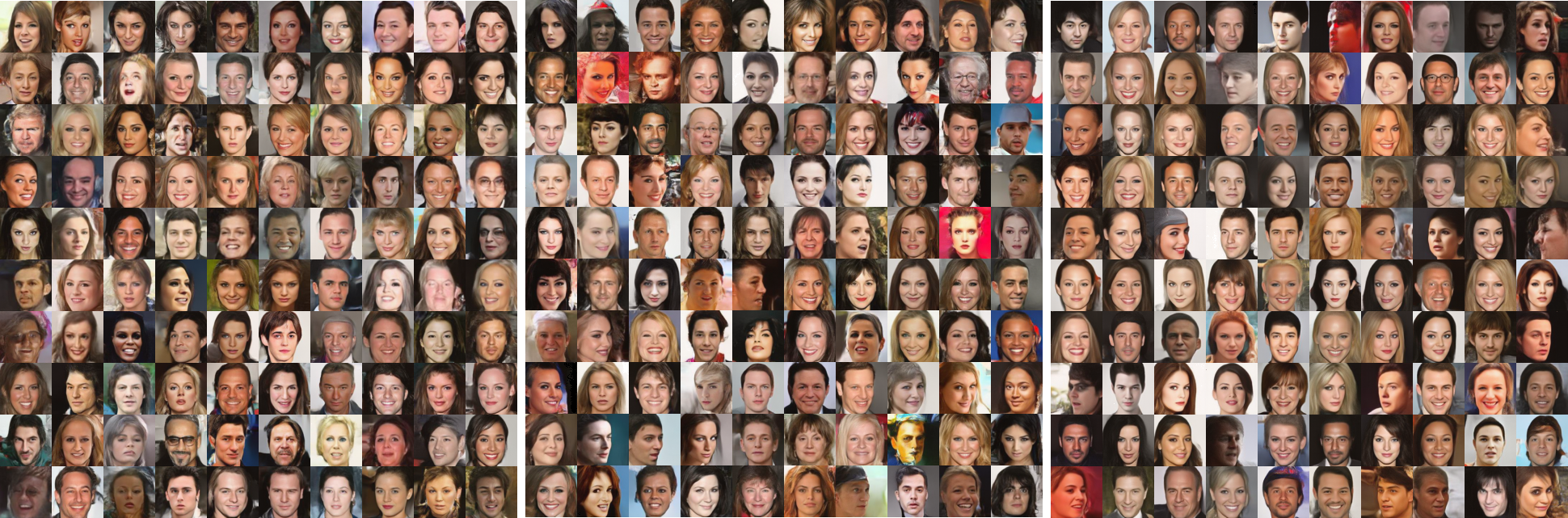}
    \caption{Random generated images from {\prior} trained with NVAE on  Celeba64 dataset. Samples are generated with temperature $t=1.0, t=0.7, t=0.4$ (from left to right).}
    \label{fig:gen_celeba_app} 
  \end{figure*}
  
\begin{figure*}[hbt] 
    \centering
        \includegraphics[width=\textwidth]{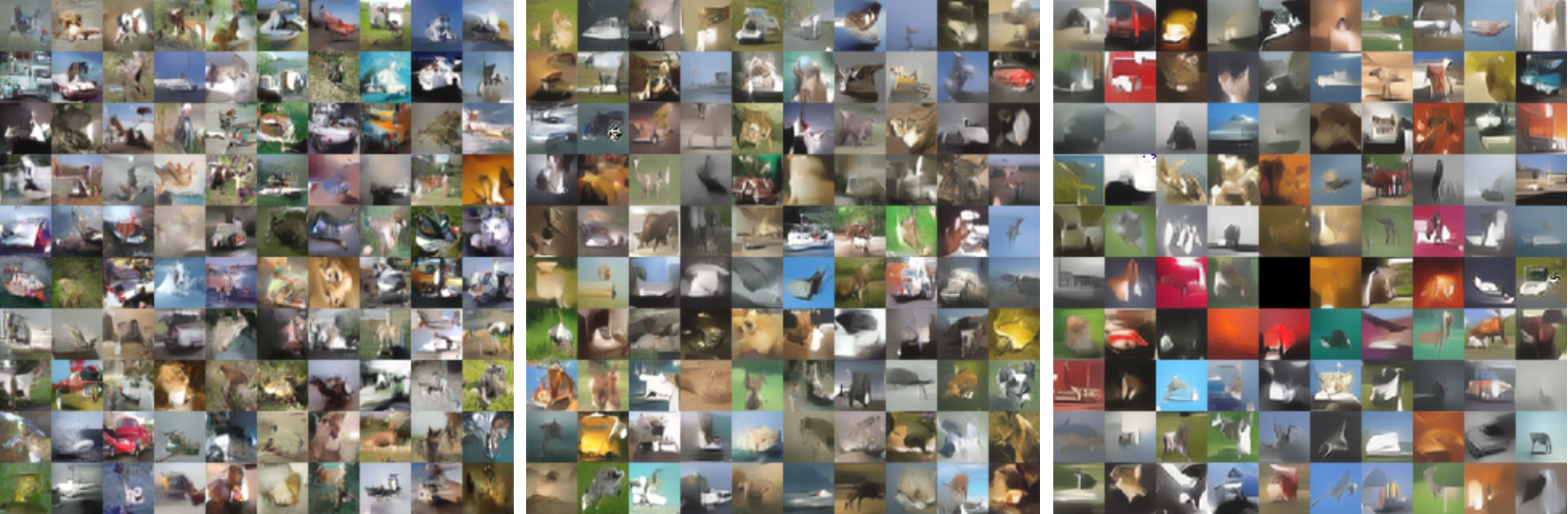}
    \caption{Random generated images from {\prior} trained with NVAE on CIFAR10 dataset. Samples are generated with temperature $t=1.0, t=0.7, t=0.4$ (from left to right).}
    \label{fig:gen_cifar_app} 
  \end{figure*}
\subsubsection{Nearest Neighbours from Trainset}
We include some more results from the nearest neighbours experiment. We follow the same method as described in section \ref{sec:nn}.  Nearest neighbours are fetched from the train set by KDTree algorithm after projecting both generated and trainset images into a $100$-dimensional PCA plane. We use Euclidean distance as a neighbour distance metric. Nearest neighbour images from single latent variable model is included in Figure \ref{fig:nn_single_latent}. Figure \ref{fig:nn_hvae} contains nearest neighbour images from the HVAE {\snis} model. Here also, we can observe that nearest neighbours from the train-set are quite different from the {\snis} generated images in both the settings (i.e. single latent variable and hierarchical latent variable VAE). This suggests that, {\snis} doesn't overfit to the train data and the gain in FID scores in Table \ref{tab:main_result_1} and \ref{tab:main_result_2} are coming from its enhanced generative capabilities. 
\vspace{-1pt}
\subsubsection{Generated Images}
 We include some more generated images for qualitative assessment from the trained models on VAE with single latent group as well with multiple latent groups. 
 Figure \ref{fig:gen_small_app} contains images from {\snis} trained on single latent group VAEs. Refer to Table \ref{tab:main_result_1} for quantitative performance of these models. \\
 
 Figure \ref{fig:gen_celeba_app} includes generated samples from the {\snis} model on Celeba64 and Figure \ref{fig:gen_cifar_app} includes generated samples from {\snis} trained on CIFAR10 (see Table \ref{tab:main_result_2} for quantitative results).  


\end{document}